\documentclass[10pt,twocolumn,letterpaper]{article}

\usepackage{cvpr}              

\usepackage{graphicx}
\usepackage{amsmath, bm}
\usepackage{amssymb}
\usepackage{booktabs}
\usepackage{amsthm}
\usepackage{colortbl}
\usepackage{comment}
\usepackage[dvipsnames]{xcolor}
\usepackage[pagebackref,breaklinks,colorlinks]{hyperref}
\usepackage{color}

\usepackage[capitalize]{cleveref}

\usepackage{hyperref}

\crefname{section}{Sec.}{Secs.}
\Crefname{section}{Section}{Sections}
\Crefname{table}{Table}{Tables}
\crefname{table}{Tab.}{Tabs.}

\def\cvprPaperID{4556} 
\def\confName{CVPR}
\def\confYear{2022}

\begin{document}

\title{Discrete Point-wise Attack Is Not Enough: Generalized Manifold Adversarial Attack for Face Recognition}

\author{Qian li, Yuxiao Hu\thanks{Co-first author}, Ye Liu, Dongxiao Zhang, Xin Jin\thanks{Corresponding author}, Yuntian Chen\footnotemark[2]\\
Eastern Institute for Advanced Study, Ningbo Zhejiang, China\\
{\tt\small \{QianL01205, huyuxiao1205, liuye66a\}@gmail.com
}\\
{\tt\small \{dzhang, jinxin, ychen\}@eias.ac.cn
}
}

\maketitle



\newtheorem{theorem}{Theorem}[]
\newtheorem{definition}{Definition}[]
\newtheorem{remark}{Remark}[]

\begin{abstract}

    Classical adversarial attacks for Face Recognition (FR) models typically generate discrete examples for target identity with a single state image. However, such paradigm of point-wise attack exhibits poor generalization against numerous unknown states of identity and can be easily defended. In this paper, by rethinking the inherent relationship between the face of target identity and its variants, we introduce a new pipeline of Generalized Manifold Adversarial Attack (GMAA)\footnote{\url{https://github.com/tokaka22/GMAA}} to achieve a better attack performance by expanding the attack range. Specifically, this expansion lies on two aspects -- GMAA not only expands the target to be attacked from one to many to encourage a good generalization ability for the generated adversarial examples, but it also expands the latter from discrete points to manifold by leveraging the domain knowledge that face expression change can be continuous, which enhances the attack effect as a data augmentation mechanism did. Moreover, we further design a dual supervision with local and global constraints as a minor contribution to improve the visual quality of the generated adversarial examples. We demonstrate the effectiveness of our method based on extensive experiments, and reveal that GMAA promises a semantic continuous adversarial space with a higher generalization ability and visual quality.

\end{abstract}

\vspace{-1em}
\section{Introduction}
\vspace{-0.5em}
\label{intro}
\begin{figure*}
\centering
\includegraphics[width=0.98\textwidth]{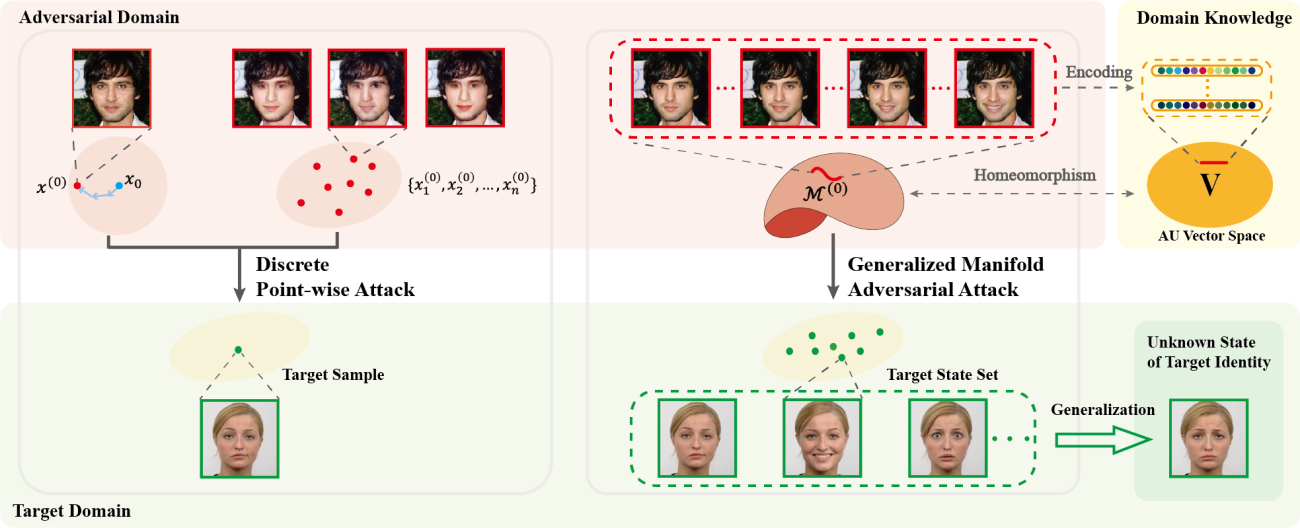}
\caption{The core idea comparison. Discrete point-wise attack methods leverage a single state of the target identity during training and provide discrete adversarial examples. By attacking the target identity's state set and employing domain knowledge, our core idea aims to Generalized Manifold Adversarial Attack (GMAA), which promises a semantic continuous adversarial manifold with a higher generalization ability on the target identity with unknown state. }
\vspace{-0.5em}
\label{intro2}
\end{figure*}
Thanks to the rapid development of deep neural networks (DNNs), the face recognition (FR) networks \cite{facenet,mb} has been applied to identity security identification systems in a variety of crucial applications, such as face unlocking and face payment on smart devices. However, it has been observed that DNNs are easily fooled by adversarial examples and offer incorrect assessments \cite{pgd,dong19}, which has risks of unauthorized access to FR systems and stealing personal privacy through poisoned data. These `well-designed' adversarial examples reveal the aspects of FR models that are vulnerable to be attacked, which makes the adversarial attack a meaningful work to provide reference for improving model robustness.

\begin{figure}
\centering
\includegraphics[width=0.45\textwidth]{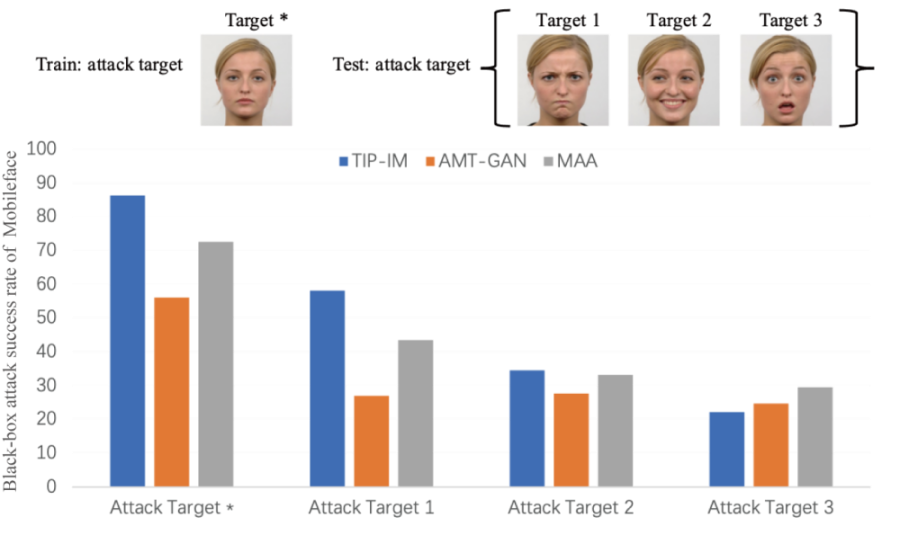}
\vspace{-1em}
\caption{The black-box attack success rate on the Mobileface of attacking target * 1, 2 and 3 during the testing. The three methods are exclusively trained on target *.}
\vspace{-1em}
\label{intro3}
\end{figure}

Following the point-wise paradigm, previous methods typically tend to attack a single target identity sample with discrete adversarial examples illustrated in Fig. \ref{intro2}. However, these methods are not strong enough both in target domain and adversarial domain. Concretely, for the target domain, attacking a single identity image has a poor generalization on those unseen faces (even if they belong to the same person) in realistic scenarios. For example, Fig. \ref{intro3} shows these adversarial examples which were used to attack the \emph{target} (a girl) have a disappointingly lower success rate of attacking the identity with other unseen states. We analyze that's because attacking a single image of target identity overfits the generation of adversarial examples to some fixed factors such as expression, makeup style, etc. In addition to the target domain, we naturally consider the weakness of adversarial domain in the existing methods. Most adversarial attack methods optimize a $L_p$ bounded perturbation~\cite{madry17,zj21iccv} based on the gradient, which limits the problem to searching for discrete adversarial examples in a hypersphere of the clean sample and ignores the continuity of the generated adversarial domain. For example, the recently proposed methods based on makeup style transformation~\cite{icip19,advmakeup,AMT} focus on generating finite adversarial examples that correspond to discrete makeup references. Such methods all overemphasize on mining discrete adversarial examples within a limited scope and ignore the importance of the continuity in adversarial space. The weakness of current adversarial attack tasks motivates us, on the one hand, to explore how to generate adversarial examples that are more general to various target identity's states and, on the other hand, to upgrade adversarial domain from discrete points to continuous manifold for a stronger attack.

In this paper, we introduce a new paradigm dubbed Generalized Manifold Adversarial Attack (GMAA) depicted in Fig. \ref{intro2}, which achieves a higher attack success rate by providing semantic continuous adversarial domains while expanding
the target to be attacked from an instance to a group set. Specifically, we train adversarial examples to attack a target identity set rather than a single image, which increases the attack success rate on the target identity with different states. The expansion of target domain naturally prompts us to consider enhancing the adversarial domain. Inspired by the success of some recent \emph{data and knowledge dual driven methods}~\cite{chen22,kar21,ZhangB}, we explore a low-dimensional manifold near the sample according to the domain knowledge, which is a simple yet highly-efficient continuous embedding scheme and can be used to augment the data. For such manifold, the data in it share the visual identity same as the original sample and also lie in the decision boundaries of target identity. More specifically, we employ the Facial Action Coding System (FACS)~\cite{FACS} as prior domain knowledge (a kind of instantiation) to expand adversarial domain from discrete points to manifold. Through using FACS, the \emph{expressions} can be encoded into a vector space, and by which the adversarial example generator could produce an manifold that is homogeneous with the expression vector space and possesses semantic continuity. In addition, in order to build an adversarial space with high visual quality, we employ four expression editors in GMAA pipeline to supervise the adversarial example generation in terms of global structure and local texture details. A transferability enhancement module is also introduced to drive the model to mine robust and transferable adversarial features. Extensive experiments have shown that these components work well on a wide range of baselines and black-box attack models.

Our contributions are summarized as follows.
\vspace{-1em}
\begin{itemize}
  \item [$\bullet$] We first pinpoint that the popular adversarial attack methods generally face generalization difficulty caused by the limited point-wise attack mechanism.
  \vspace{-0.8em}
  \item [$\bullet$] To enhance the performance of adversarial attack, a new paradigm of Generalized Manifold Adversarial Attack (GMAA) is proposed with an improved attack success rate and better generalization ability.
  \vspace{-0.8em}
  \item [$\bullet$] GMAA considers the enhancement in terms of both target domain and adversarial domain. For the target domain, it expands the target to be attacked from one to many to encourage a good generalization. For the adversarial domain, the domain knowledge is embedded to strengthen the attack effect from discrete points to continuous manifold.
  \vspace{-0.8em}
  \item [$\bullet$] We instantiate GMAA in the face expression state space for a semantic continuous adversarial manifold and use it to attack a state set of the target identity. As a minor contribution, GMAA supervises the adversarial example generator w.r.t global structure and local details with the pre-trained expression editors for a high visual quality.
\end{itemize}


\vspace{-1em}
\section{Related work}
\vspace{-0.5em}
\subsection{Adversarial Attacks on Face Recognition}
\vspace{-0.5em}
Adversarial attacks can be divided into white-box attacks and black-box attacks. White-box attacks \cite{Ehae, making20, madry17} must entail full information of the target neural networks, however, the parameters and architecture of a real-world face recognition system are typically hard to access. Hence, it is more practical to consider black-box adversarial attacks in face recognition scenarios, which demand high transferability for the unknown FR target models or commercial application programming interfaces (APIs). Query-based adversarial attacks \cite{dong19, YingGuo}, a form of black-box attack, is inappropriate for realistic applications since it optimizes adversarial examples by repeatedly accessing the target model during training. Adding transferable adversarial perturbation is another effective black-box attack form \cite{boosting, zj21iccv, evading, advface}. Unfortunately, the perturbation caused by this method always makes images become unrealistic and unnatural. Besides, limited by the way of pixel-to-pixel similarity computation, the model of this branch could only look for a limited number of discrete adversarial examples around the clean sample. The patch-based adversarial attack~\cite{advhat, xiao21, advglass} severely degrades the visual quality of images, because the adversarial patch does not blend well with the clean background due to the abrupt shift in pixel values around the boundary. Numerous recent works attempt to instantiate adversarial perturbations with different makeup styles~\cite{AMT, advmakeup, icip19}. However, such makeup attacks typically result in an unnatural visual appearance due to gender constraints -- female images have a higher attack success rate and visual quality than male ones. Regardless of the previous methods (e.g. \cite{semanticadv} focuses on a target sample to obtain optimal feature-map interpolation, and \cite{AMT} generates adversarial examples with various makeup styles to attack the target identity), they all tend to generate discrete adversarial examples for a single target identity sample and ignore the importance of continuity of adversarial space, which might be crucial to the attack performance.


\subsection{Facial Expression Editing}
\vspace{-0.5em}
As another challenging task in facial analysis, face expression editing is also related to our study, which aims at modifying facial expressions in a reasonable manner while preserving identity completeness. In recent years, generative adversarial networks (GANs) have achieved surprising advances in facial expression editing: GCGAN~\cite{GCGAN} uses the facial geometry as prior knowledge to guide the generation. ExprGAN~\cite{ExprGAN} exploits a controller to adjust the intensity of face expression editing. StarGAN~\cite{stargan} introduces a cycle consistency loss to maintain the identity content invariant. However, these methods are all limited on the discrete expressions generation. GANimation~\cite{ganimation} utilizes \emph{Action Units}~\cite{FACS} to define an expression space and generate continuous-change facial images. EF-GAN~\cite{EF} achieves progressive facial expression editing with a local-focused cascade GAN structure, and produces fewer artifacts and blurs in large-gap expression transformations. In this paper, we borrow the idea of \emph{Action Units} as domain prior knowledge (from face expression editing) to re-define the adversarial attack task on a continuous manifold, to finally strengthen attack effect.

\vspace{-0.5em}
\section{Generalized Manifold Adversarial Attack}
\begin{figure*}
\centering
\includegraphics[width=0.95\textwidth]{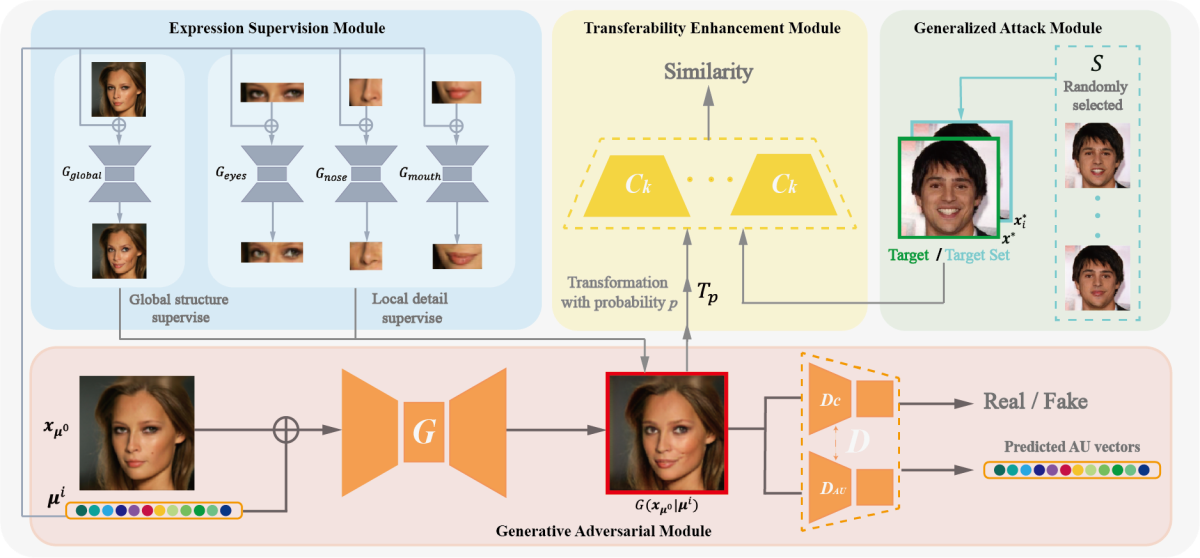}
\vspace{-0.5em}
\caption{Overview of GMAA. A face image and a random AU vector are sent to the generative adversarial module as inputs. Meanwhile, the expression supervision module generates supervisory signals based on the AU vector and the input image (or detail patches cropped by the landmarks). In the transferability enhancement module, aiming at white-box FR models, the transformed output attacks the target identity's state set, which is provided by the generalized attack module.}
\vspace{-1.5em}
\label{structure}
\end{figure*}
\subsection{Problem Definition}
Adversarial face attack tasks can be separated into targeted attacks (i.e. impersonation attacks) and non-targeted attacks (i.e. dodging attacks). Targeted attacks force the generated adversarial examples to have a predetermined output towards the target FR model, while the non-targeted attacks mislead the target FR model to provide incorrect random classification for the adversarial examples. To capture the adversarial examples that can impersonate a specific target identity under face authentication systems, we mainly consider the targeted attack task.

Current methods always define the targeted attack task as an optimization problem, which can be formalized as
\vspace{-0.3em}
\begin{equation}
\begin{aligned}
\mathop{\rm min}L_{adv}&=\mathop{\rm min}\limits_{\bm{\theta}}Dist(\mathcal{C}(\bm{x}^{*}),\mathcal{C}(\bm{x}')),\\
\bm{x}'&=G(\bm{x};\bm{\theta}).
\end{aligned}
\vspace{-0.3em}
\end{equation}
where $\bm{x}^{*}$ is the pre-specified target image belonging to the sample space $\Omega\subset \mathbb{R}^{3\times H\times W}$, $Dist(\cdot)$ represents a metric of difference, $\mathcal{C}$ represents the feature extractor of FR neural networks, and $G$ maps the clean sample $\bm{x}\in\Omega$ to the adversarial version $\bm{x}'$ with the parameter $\bm{\theta}$.

In this paper, we re-define the problem from a broader standpoint. To obtain highly generalized adversarial examples that are more threatening to the target identity with unknown state, the adversarial version $\bm{x}'$ attacks the state set $\mathcal{S}$ of the target identity during training. In order to capture an adversarial manifold $\mathcal{M}$ instead of discrete adversarial examples, we aim to construct a distribution on the $\mathcal{M}$. The new task can be formalized as Eq. \ref{ntask}.
\vspace{-0.3em}
\begin{equation}
\begin{aligned}\label{ntask}
\mathop{\rm min}L_{adv}&=\mathop{\rm min}\limits_{\bm{\theta}}\mathbb{E}_{\bm{x}_i^*\sim\mathcal{S},\bm{x}'\sim\mathcal{M}}Dist(\mathcal{C}(\bm{x}_i^{*}),\mathcal{C}(\bm{x}')),\\
\mathcal{M}&=G(\bm{x};\bm{\theta}).
\end{aligned}
\end{equation}

In addition to the adversarial attack tasks, we force $G$ to map $\bm{x}$ to $\bm{x}'$ according to the given expression AU vector. The Facial Action Coding System (FACS) is applied as prior domain knowledge to establish a continuous expression state space. Specifically, every expression is encoded by an n-element expression vector $\bm{\mu}=(\mu_1,...,\mu_n)$ which corresponds to $N$ facial action units. Each $\mu_n\in\bm{\mu}$ represents the magnitude of muscle activity in the n-th region of the face, which indicates that the AU vector is a continuous embedding scheme for different expressions. Thus, for the input image ${\bm x}_{{\bm\mu}^0}\in\mathbb{R}^{3\times H\times W}$ with the expression encoded by ${\bm\mu}^0\in\mathbb{R}^{N}$ and the given expression encoded by ${\bm\mu}^i\in\mathbb{R}^{N}$, $G$ is a binary mapping $G:({\bm x}_{{\bm\mu}^0},\bm{\mu}^i)\to {\bm x}_{{\bm\mu}^i}'$, where ${\bm x}_{{\bm\mu}^i}'$ has the same visual label with ${\bm x}_{{\bm\mu}^0}$ and wearing the expression encoded by $\bm{\mu}^i$. 

\vspace{-0.5em}
\subsection{Generalized Manifold Adversarial Attack}
\vspace{-0.5em}
We establish the adversarial examples' distribution on a manifold via WGAN-GP \cite{wgangp}. GMAA comprises a generative adversarial module, an expression supervision module, a transferability enhancement module, and a generalized attack module. In contrast, Manifold Adversarial Attack (MAA) omits the generalized attack module, as it merely extends the adversarial domain from a point to a manifold. The structure of our proposed method is depicted in Fig. \ref{structure}.

\noindent{\bf Generative adversarial module.}
The generative adversarial module (red box in Fig. \ref{structure}) includes a generator $G$, a discriminator $D_c$ and an AU predictor $D_{AU}$, where $D_c$ and $D_{AU}$ both lie in $D$ and share partial parameters. 

As inputs, the generator $G$ receives a clean sample ${\bm x}_{{\bm\mu}^0}$ and a given expression AU label ${\bm\mu}^i$, which aims to produce adversarial examples wearing the expression matching to the supplied AU label and maintain the same visual identity with ${\bm x}_{{\bm\mu}^0}$. The discriminator $D_c$ learns to distinguish real images from generated images. Meanwhile, generated images deceive the discriminator $D_c$ to force the outputs of the generator $G$ match the real distribution. Our $G$ and $D_c$ are trained using WGAN-GP \cite{wgangp}, and the critic loss function we employ is 
\vspace{-0.3em}
\begin{equation}
\begin{aligned}
L_{critic}^{D}=&\lambda_{c}(1-D_c({\bm x}_{{\bm\mu}^0}))^2+\lambda_{c}(D_{c}(G({\bm x}_{{\bm\mu}^0}|{\bm\mu}^i))^2\\
&+\lambda_{gp}(\lVert\nabla_{\widetilde{\bm x}}D_c(\widetilde{\bm x})\rVert_2-1)^2,
\end{aligned}
\end{equation}
\vspace{-1em}
\begin{equation}
\begin{aligned}
L_{critic}^G=\lambda_{c}(1-D_c(G({\bm x}_{{\bm\mu}^0}|{\bm\mu}^i)))^2,
\end{aligned}
\end{equation}
where $\widetilde{\bm x}$ is the random interpolation distribution between the real distribution and the generated images' distribution. To ensure that the generated image match the provided expression code ${\bm\mu}^i$, we employ AU regression loss to establish the consistency of the generated expression with ${\bm\mu}^i$. Specifically, the AU predictor $D_{AU}$ learns the AU coding rules by real images and their AU labels (can be obtained by the open source framework Openface \cite{openface}), and the $G$ reduces the AU error between the generated expression and ${\bm\mu}^i$ to satisfy the given expression. The loss function can be formulated as: 
\vspace{-0.3em}
\begin{equation}
\begin{aligned}
&L_{AU}^D=\lambda_{AU}\lVert D_{AU}({\bm x}_{{\bm\mu}^0})-{\bm\mu}^0\lVert_2^2,\\
&L_{AU}^G=\lambda_{AU}\lVert D_{AU}(G({\bm x}_{{\bm\mu}^0}|{\bm\mu}^i))-{\bm\mu}^i\rVert_2^2.
\end{aligned}
\end{equation}
\vspace{-1em}

\noindent{\bf Expression supervision module.}
The expression supervision module (blue box in Fig. \ref{structure}) protects the visual identity of adversarial examples and guides $G$ in expression editing by generating global and local facial supervisory signals. The global branch focuses on structural features of the face, whereas the local branch protects important facial details and reduces artifacts and blurs caused by the global branch.

Specifically, a global editor and three local editors are pre-trained to provide supervisory signals. For the local editors, we crop the eyes, nose, and mouth pixel patches based on face landmarks first. Then the input image and three detail patches are fed into the corresponding generator $G_{global}$ and $G_j$ ($j\in J=$\{eyes, nose, mouth\}) with the input AU vector ${\bm\mu}$, respectively. Each generator has the network structure similar to \cite{ganimation}, which provides the color response $M_c^{\bm\mu}$ and the attention response $M_a^{\bm\mu}$. $M_c^{\bm\mu}$ and $M_a^{\bm\mu}$ force networks to pay attention to the expression change region and protect the remainder regions from disturbance. The ultimate supervisory signals can be obtained as follows, where $\otimes$ denotes element-wise multiplication.
\vspace{-0.5em}
\begin{equation}
\begin{aligned}
G_j({\bm x}_{in}|{\bm\mu})=M_a^{\bm\mu}\otimes M_c^{\bm\mu}+(1-M_a^{\bm\mu})\otimes{\bm x}_{in}.
\end{aligned}
\end{equation}
\vspace{-1.5em}

The global editor focuses more on large scale features, such as shape and position of the five senses, and tends to produce artifacts and blurs in the detail region, while the local editors concentrate on significant local features and provide finer details. Therefore, we deploy the global editor to supervise the structural information of adversarial examples and the local editors to supervise local specifics. The loss of expression supervision module can be expressed as
\begin{equation}
\begin{aligned}
&L_{exp}^G=\lambda_{g}{\rm SSIM}[G({\bm x}_{{\bm\mu}^0}|{\bm\mu}^i),G_{global}({\bm x}_{{\bm\mu}^0}|{\bm\mu}^i)]\\
&+\lambda_{l}\sum_{j\in J}{\rm MSE}[{\rm Crop}_j(G({\bm x}_{{\bm\mu}^0}|{\bm\mu}^i)),G_j({\rm Crop}_j({\bm x}_{{\bm\mu}^0})|{\bm\mu}^i)],\\
\end{aligned}\label{exploss}
\end{equation}
where ${\rm Crop}_j$ denotes the crop operation of local region $j$ according to the face landmarks.

\noindent{\bf Transferability enhancement module.} 
To improve the transferability of adversarial examples and the black-box attack success rate, we introduce the transferability enhancement module (yellow box in Fig. \ref{structure}) from \cite{AMT}. The generated adversarial examples are transformed with probability $p$ by the function $T_p$ (resize with padding or add noise), and then attack $K$ pre-trained high-precision FR models $\{C_k\}_{k=1,...K}$, which perform as white-box models during the training of $G$. The adversarial attack loss function is formulated as
\vspace{-0.5em}
\begin{equation}\label{advloss}
\begin{aligned}
&L_{adv}^G=\frac{\lambda_{adv}}{K}\sum_{k=1}^K[1-\cos(\mathcal{C}_k({\bm x}^*),\mathcal{C}_k(T_p(G({\bm x}_{{\bm\mu}^0}|{\bm\mu}^i))))]\\
\end{aligned}
\end{equation}

\noindent{\bf Generalized attack module.} 
The generalized attack module (green box in Fig. \ref{structure}) intends to raise the attack success rate on the unseen face belonging to the target identity, which can be introduced into other adversarial attack approaches. The adversarial loss $L_{adv}^G$ in GMAA can be further expressed as
\vspace{-0.5em}
\begin{equation}
\begin{aligned}
\mathbb{E}_{\bm{x}_i^*\sim\mathcal{S}}\frac{\lambda_{adv}}{K}\sum_{k=1}^K[1-\cos(\mathcal{C}_k({\bm x}^*_i),\mathcal{C}_k(T_p(G({\bm x}_{{\bm\mu}^0}|{\bm\mu}^i))))],
\end{aligned}
\end{equation}
where ${\bm x}^*_i$ and $\mathcal{S}$ are defined in \ref{ntask}. However, numerous face recognition datasets and realistic scenarios do not fit this module since an identity only contain a few or single state image. Fortunately, since our method can accomplish both expression editing and adversarial attack, when we remove the loss associated with adversarial attack, we can obtain an expression editor, $G_{exp}$, by eliminating the adversarial effect $L^G_{adv}$, which can generate the expression state set $\mathcal{S}$ by different AU vectors.

\noindent{\bf Total loss function.}
Let ${\bm X}$ denotes the dataset, and $V$ is the AU vector space. In particular, the given AU vector ${\bm\mu}^i$ is sampled randomly from $V$ to train the generator $G$ to learn the distribution of adversarial expression manifold. For the generator $G$, we have the loss function as follows, 
\vspace{-0.5em}
\begin{equation}\label{totalg1}
\begin{aligned}
L^G=\mathbb{E}_{{\bm x}_{{\bm\mu}^0}\sim{\bm X},{{\bm\mu}^i\sim V}}
(L^G_{critic}+L^G_{AU}+L^G_{exp}+L^G_{adv}).
\end{aligned}
\end{equation}
As for $D$, the total loss fuction can be obtained as follows,
\vspace{-0.5em}
\begin{equation}\label{totald}
\begin{aligned}
L^D=\mathbb{E}_{{\bm x}_{{\bm\mu}^0}\sim{\bm X},{{\bm\mu}^i\sim V}}
(L^D_{critic}+L^D_{AU}).
\end{aligned}
\end{equation}
\begin{table*}
\caption{Black-box attack success rate}
\vspace{-0.8em}
\centering
\footnotesize
\setlength{\tabcolsep}{2.5mm}{
\begin{tabular}{ccccccccc}
\toprule
&\multicolumn{4}{c}{\textbf{\underline{\qquad\qquad\qquad CelebA-HQ\qquad\qquad\qquad}}}& \multicolumn{4}{c}{\textbf{\underline{\qquad\qquad\qquad\qquad LFW\qquad\qquad\qquad\qquad}}}\\
&IRSE50&IR152&Facenet&Mobileface&IRSE50&IR152&Facenet&Mobileface\\
\midrule
Clean&3.68&3.08&1.31&8.43&3.20&0.06&0.04&5.00\\
\midrule
PGD\cite{pgd}&24.20&13.37&5.86&28.72&31.30&10.20&7.40&33.50\\
MI-FGSM\cite{boosting}&38.90&\cellcolor{Yellow!30} 20.76&9.25&40.48&38.20&14.20&7.60&39.40\\
SemanticAdv\cite{semanticadv}&26.53&10.24&7.80&55.32&33.60&10.40&8.80&37.40\\
TIP-IM\cite{zj21iccv}&44.20&16.09&\cellcolor{Yellow!30}14.46&\cellcolor{Yellow!80}\textbf{65.36}&32.80&15.20&13.00&\cellcolor{Yellow!80}\textbf{79.00}\\
AMT-GAN\cite{AMT}&\cellcolor{Yellow!30}51.06&15.63&11.63&33.27&\cellcolor{Yellow!30}40.72&\cellcolor{Yellow!30}25.23&\cellcolor{Yellow!30}13.89&35.67\\
\midrule
\textbf{MAA}&\cellcolor{Yellow!80}\textbf{60.40}&\cellcolor{Yellow!80}\textbf{29.43}&\cellcolor{Yellow!80}\textbf{18.91}&\cellcolor{Yellow!30}56.13&\cellcolor{Yellow!80}\textbf{55.80}&\cellcolor{Yellow!80}\textbf{29.20}&\cellcolor{Yellow!80}\textbf{18.00}&\cellcolor{Yellow!30}60.80\\
\bottomrule
\label{tsingle}
\vspace{-1.8em}
\end{tabular}}
\end{table*}

\begin{table*}
\caption{The unbolded numbers are the black-box ASR of attacking the test target *, 1, 2, 3, and all the models are trained on target *. The bolded numbers are the results of the models that train on the state set. For example, if we train the MAA to attack the state set without target 1, the adversarial examples attack target 1 with an 11.43\% success rate on Facenet during the testing period.}
\vspace{-0.8em}
\centering
\footnotesize
\setlength{\tabcolsep}{1.4mm}{
\begin{tabular}{ccccccccc}
\toprule
&\multicolumn{2}{c}{\textbf{\underline{\qquad\quad Target*\quad\qquad}}}& \multicolumn{2}{c}{\textbf{\underline{\quad\qquad Target 1\quad\qquad}}}& \multicolumn{2}{c}{\textbf{\underline{\quad\qquad Target 2\quad\qquad}}}& \multicolumn{2}{c}{\textbf{\underline{\quad\qquad Target 3\quad\qquad}}}\\
&Facenet&Mobileface&Facenet&Mobileface&Facenet&Mobileface&Facenet&Mobileface\\
\midrule
TIP-IM\cite{zj21iccv}~/~\textbf{G-TIP-IM} &17.68&86.33&4.54~/~\textbf{7.62}&58.03~/~\textbf{70.93}&10.75~/~\textbf{20.42}&34.42~/~\textbf{49.20}&11.93~/~\textbf{19.41}&22.21~/~\textbf{42.43}\\
AMT-GAN\cite{AMT}~/~\textbf{G-AMT-GAN} &16.12&55.95&8.22~/~\textbf{13.23}&26.99~/~\textbf{47.14}&9.78~/~\textbf{17.12}&27.67~/~\textbf{43.93}&10.91~/~\textbf{16.16}&24.69~/~\textbf{42.37}\\
MAA~/~\textbf{GMAA}&25.22&72.62&11.43~/~\textbf{17.84}&43.44~/~\textbf{67.50}&13.30~/~\textbf{21.71}&33.08~/~\textbf{41.24}&12.64~/~\textbf{19.15}&29.56~/~\textbf{47.21} \\
\bottomrule
\label{rss}
\vspace{-2.3em}
\end{tabular}}
\end{table*}

\begin{table*}
\caption{The effect of the image quantity in state set on improving the generalizability of the adversarial example. The values ${n_i}$ in ${n_1/n_2/n_3}$ represents each expression state using ${i}$ images.}
\vspace{-0.5em}
\centering
\footnotesize
\setlength{\tabcolsep}{3.1mm}{
\begin{tabular}{ccccccccc}
\toprule
&\multicolumn{2}{c}{\textbf{\underline{\quad\qquad Target 1\quad\qquad}}}& \multicolumn{2}{c}{\textbf{\underline{\quad\qquad Target 2\quad\qquad}}}& \multicolumn{2}{c}{\textbf{\underline{\quad\qquad Target 3\quad\qquad}}}\\
&Facenet&Mobileface&Facenet&Mobileface&Facenet&Mobileface\\
\midrule
G-TIP-IM \cite{zj21iccv}&7.6\,/\,7.1\,/\,7.2&70.9\,/\,70.3\,/\,70.1&20.4\,/\,21.5\,/\,21.4&49.2\,/\,49.2\,/\,49.2&19.4\,/\,19.6\,/\,19.4&39.4\,/\,39.6\,/\,39.5\\
G-AMT-GAN \cite{AMT}&13.2\,/\,14.4\,/\,13.9&47.1\,/\,44.8\,/\,45.3&17.1\,/\,16.9\,/\,18.9&43.9\,/\,45.4\,/\,43.8&16.2\,/\,15.4\,/\,15.7&42.4\,/\,44.6\,/\,45.4\\
GMAA&17.8\,/\,18.2\,/\,17.7&67.5\,/\,68.5\,/\,69.3&21.7\,/\,21.3\,/\,19.6&41.2\,/\,42.6\,/\,40.3&19.2\,/\,19.9\,/\,21.3&47.2\,/\,45.6\,/\,44.7 \\
\bottomrule
\label{rss2}
\vspace{-1em}
\end{tabular}}
\end{table*}
\vspace{-0.5em}
\begin{figure*}
\centering
\includegraphics[width=1\textwidth]{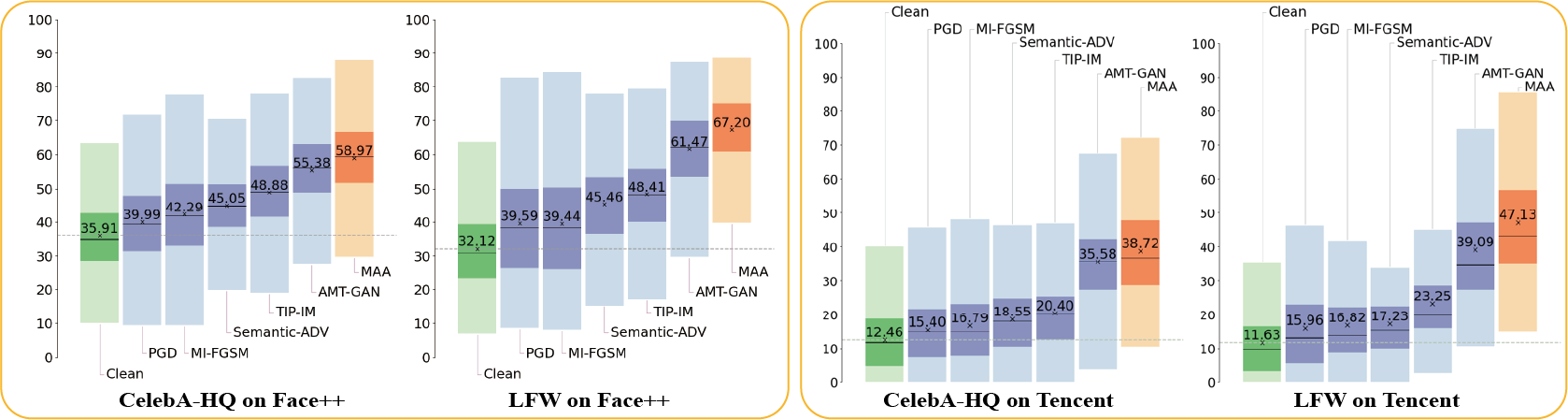}
\vspace*{-1.5em}
\caption{The confidence scores returned by Face++ and Tencent. The dashed line represents the average confidence level of clean samples.}
\vspace{-1em}
\label{api}
\end{figure*}

\begin{figure}
\centering
\includegraphics[width=0.48\textwidth]{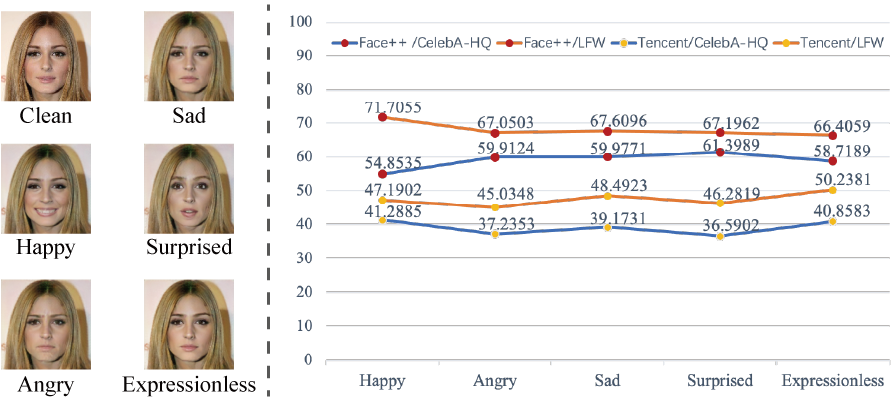}
\caption{The left side of the image depicts the five expression states, while the right side of the image depicts the influence of varied AU on the attack performance of Face++ and Tencent.}
\vspace*{-0.5em}
\label{apiexp}
\end{figure}
\begin{figure*}
\centering
\includegraphics[width=1\textwidth]{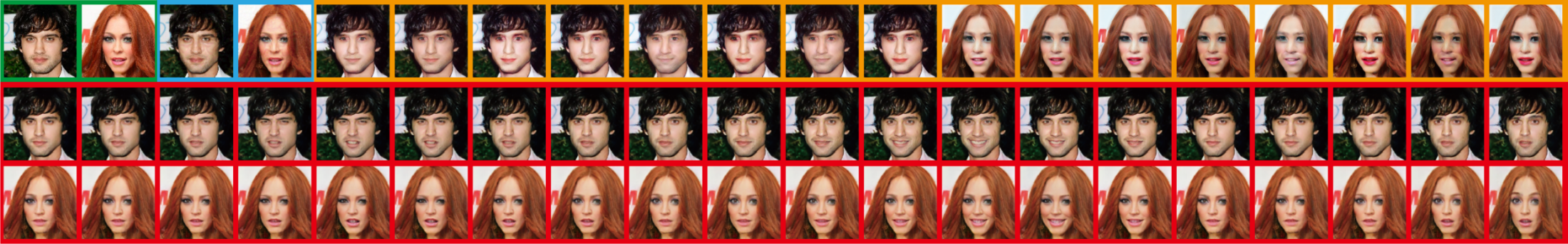}
\vspace*{-1.5em}
\caption{The images with green frames are the clean samples, while the images with blue frames are the results of TIP-IM \cite{zj21iccv}. In the case of the AMT-GAN \cite{AMT}, we chose 8 makeup styles at random, and the visualization results are shown as images with an orange frame. The images highlighted by red frames are the results of MAA, which are generated by a set of AU vectors. Please refer to the supplementary material for more high-definition magnified visualization results.}
\vspace{-1em}
\label{vis}
\end{figure*}

\begin{table}
\caption{This table shows the black-box ASR results of attacking the test target, which is the same person as the train image highlighted by a green square in Fig. \ref{structure}. The unbolded numbers represent the results of training on the single target image that is shown in Fig. \ref{structure}, while the bolded numbers are the results of training on the generated state set.}
\vspace{-1em}
\centering
\footnotesize
\setlength{\tabcolsep}{3mm}{
\begin{tabular}{ccccccccc}
\toprule
&Facenet&Mobileface\\
\midrule
TIP-IM\cite{zj21iccv}~/~\textbf{G-TIP-IM}&5.80~/~\textbf{9.50}&17.20~/~\textbf{23.5}\\
AMT-GAN\cite{AMT}~/~\textbf{G-AMT-GAN}&4.04~/~\textbf{8.27}&9.82~/~\textbf{12.45}\\
MAA~/~\textbf{GMAA}&6.60~/~\textbf{10.60}&13.50~/~\textbf{21.60}\\
\bottomrule
\label{gss}
\vspace{-3em}
\end{tabular}}
\end{table}

\vspace{-1.5em}
\subsection{Continuity of the adversarial space}
\vspace{-0.5em}
In this subsection, we illustrate more precisely the continuity of the adversarial space and provide a proof that our method establishes a continuous adversarial manifold.

Firstly, the definition of continuous adversarial space is shown in Def. \ref{def1}.
\begin{definition}\label{def1}
Let $\bm{x}_0\in\mathbb{R}^{3\times H\times W}$, then $\mathcal{M}^0=G(\bm{x}_0;\bm{\theta})$ is a continuous adversarial space if and only if \\
(1) $\mathcal{M}^0$ is a subspace of $\mathbb{R}^{3\times H\times W}$.\\
(2) $\forall \bm{x}_i^0\in\mathcal{M}$, $\bm{x}_i^0$ is an adversarial version of $\bm{x}_0$.
\end{definition}
\vspace{-0.5em}
Then, we can prove that $\mathcal{M}^0$ is a continuous manifold.
\vspace{-0.5em}
\begin{theorem}\label{thm1}
 $\mathcal{M}^0$ generated by $G_0$ is a continuous adversarial manifold, where $G_{0}: V\to\mathcal{M}$ is a map when fixed the input $\bm{x}_0$ in $G$.
\end{theorem}
\vspace{-1em}
\begin{proof}
(1) If $\mathcal{M}^0$ homogeneous with the AU vector space $V$, it is obviously that $\mathcal{M}^0$ is a subspace of $\mathbb{R}^{3\times H\times W}$.

(1.1) $\forall$ $\bm{\mu}$, $\bm{\nu} \in V$, if $G_{0}(\bm{\mu})=G_{0}(\bm{\nu})$, then $D_{AU}(G_{0}(\bm{\mu}))=D_{AU}(G_{0}(\bm{\nu}))$, we have $\bm{\mu}=\bm{\nu}$ and $G_{0}:V\to\mathcal{M}^{0}$ is a single shot. Besides, $\forall \bm{x}^{0}\in\mathcal{M}^{0}$, $\exists \bm{\mu}=D_{AU}(\bm{x}^{0})\in V$ s.t. $G_{0}(D_{AU}(\bm{x}^{0}))=\bm{x}^{0}$, then $G_{0}$ is a surjection. Thus, $G_{0}$ is a bijection.

(1.2) $\forall \bm{x}_1^{0}$, $\bm{x}_2^{0}\in\mathcal{M}^{0}$, we define $d(\bm{x}_1^{0}, \bm{x}_2^{0})=\lVert D_{AU}(\bm{x}_1^{0})-D_{AU}(\bm{x}_2^{0})\rVert_2$ as the metric between $\bm{x}_1^{0}$ and $\bm{x}_2^{0}$ in $\mathcal{M}^{0}$. Since $(V,\lVert \cdot\rVert_2)$ is a metric space, and $D_{AU}(\bm{x}_i^{0})\in V$, $\forall \bm{x}_i^{0}\in\mathcal{M}^{0}$, we prove that $d$ is a metric on $\mathcal{M}^{0}$.

(Positivity) $d(\bm{x}_1^{0}, \bm{x}_2^{0})\geq 0$, and if $d(\bm{x}_1^{0}, \bm{x}_2^{0})=0$, according to the definition of $d$ we have $D_{AU}(\bm{x}_1^{0})=D_{AU}(\bm{x}_2^{0})$. Since $G_{0}$ is a bijection, we have $\bm{x}_1^0=G_{0}(D_{AU}(\bm{x}_1^{0}))=G_{0}(D_{AU}(\bm{x}_2^{0}))=\bm{x}_2^0$. 

(Symmetry) $d(\bm{x}_1^{0}, \bm{x}_2^{0})=\lVert D_{AU}(\bm{x}_1^{0})-D_{AU}(\bm{x}_2^{0})\rVert_2=\lVert D_{AU}(\bm{x}_2^{0})-D_{AU}(\bm{x}_1^{0})\rVert_2=d(\bm{x}_2^{0}, \bm{x}_1^{0})$.

(Triangle inequality) $\forall \bm{x}_3^{0}\in\mathcal{M}^{0}$, we have $d(\bm{x}_1^{0}, \bm{x}_3^{0})+d(\bm{x}_3^{0}, \bm{x}_2^{0})=\lVert D_{AU}(\bm{x}_1^{0})-D_{AU}(\bm{x}_3^{0})\rVert_2+\lVert D_{AU}(\bm{x}_3^{0})-D_{AU}(\bm{x}_2^{0})\rVert_2\geq\lVert D_{AU}(\bm{x}_1^{0})-D_{AU}(\bm{x}_2^{0})\rVert_2=d(\bm{x}_1^{0}, \bm{x}_2^{0})$.

Thus, $(\mathcal{M}^{0}, d)$ is a metric space. Then we need to prove that $G_{0}$ is a continuous mapping. $\forall \bm{\mu}, \bm{\nu}\in V$, $\forall \epsilon>0$, let $\delta=\epsilon$, when $\lVert\bm{\mu}-\bm{\nu}\rVert_2<\delta$, we have $d(G_{0}(\bm{\mu})-G_{0}(\bm{\nu}))=\lVert D_{AU}(G_{0}(\bm{\mu}))-D_{AU}(G_{0}(\bm{\nu}))\rVert_2=\lVert\bm{\mu}-\bm{\nu}\rVert_2<\delta=\epsilon$. We get the continuity of $G_{0}$.

(1.3) It is obviously that $D_{AU}|_{\mathcal{M}^{0}}: \mathcal{M}^{0}\to V$ is the inverse mapping of $G_{0}$. $\forall \bm{x}_1^0, \bm{x}_2^0\in\mathcal{M}^{0}, \epsilon>0$, let $\delta=\epsilon$, when $d(\bm{x}_1^0, \bm{x}_2^0)=\lVert D_{AU}(\bm{x}_1^{0})-D_{AU}(\bm{x}_2^{0})\rVert_2<\delta$, we have $\lVert D_{AU}|_{\mathcal{M}^{0}}(\bm{x}_1^{0})-D_{AU}|_{\mathcal{M}^{0}}(\bm{x}_2^{0})\rVert_2=\lVert D_{AU}(\bm{x}_1^{0})-D_{AU}(\bm{x}_2^{0})\rVert_2<\delta=\epsilon$. We get the continuity of the inverse map of $G_{0}$.

In conclusion, $G_{0}$ is the homeomorphism of $V$ to the manifold $\mathcal{M}^{0}$, i.e. $\mathcal{M}^0$ homogeneous with the AU vector space $V$. Since AU vector space is a finite dimensional vector space, $\mathcal{M}^0$ is a subspace of $\mathbb{R}^{m\times n\times l}$.

(2) By the loss function \ref{advloss}, \ref{exploss} and the back propagation, $\bm{x}_i^0$ is influenced by the \ref{advloss}. It is obviously that $\bm{x}_i^0\in\mathcal{M}^0$ is an adversarial example when the model is well-trained.

Thus, the manifold $\mathcal{M}^0$ generated by $G_0$ is a continuous adversarial manifold.
\end{proof}
\vspace{-1.5em}
\begin{remark}
    Since the $\mathcal{M}^0$ generated by $G_0$ is a continuous adversarial manifold when fixed the $\bm{x}_0$, then we can assert over the sample space $\Omega$, the adversarial examples space generated by $G$ constitutes an adversarial fiber bundle.
\end{remark}
\vspace{-0.5em}
Secondly, the definition of semantic continuous adversarial space is shown in Def. \ref{def2}.
\vspace{-0.5em}
\begin{definition}\label{def2}
$\mathcal{M}^0$ generated by $\bm{x}_0\in\mathbb{R}^{3\times H\times W}$ is a semantic continuous adversarial space if and only if\\
(1) $\mathcal{M}^0$ is a continuous adversarial space.\\
(2) $\forall \bm{x}^0_1, \bm{x}^0_2 \in \mathcal{M}^0$, if $\bm{x}^0_1$ is close to $\bm{x}^0_2$ on the $\mathcal{M}^0$, then $\bm{x}^0_1$ and $\bm{x}^0_2$ satisfy the semantic consistency. 
\end{definition}
\vspace{-0.5em}
We can state that $\mathcal{M}^0$ is a semantic continuous manifold.
\vspace{-1.5em}
\begin{theorem}\label{thm2}
 $\mathcal{M}^0$ generated by $G_0$ is a semantic continuous adversarial manifold, where $G_{0}: V\to\mathcal{M}$ is a map when fixed the input $\bm{x}_0$ in $G$.
\end{theorem}
\vspace{-1.5em}
\begin{proof}
Since we have proved that $D_{AD}$ is a continuous mapping, we have $D_{AU}({\bm x}^0_1)$ is close to $D_{AU}({\bm x}^0_2)$ when ${\bm x}^0_1$ is close to ${\bm x}^0_2$, which means the AU vectors of ${\bm x}^0_1$ and ${\bm x}^0_2$ are very close. Thus, the semantic information of ${\bm x}^0_1$ and ${\bm x}^0_2$ is close.
\end{proof}

\vspace{-1em}
\section{Experiments}
\subsection{Experimental setting}
\vspace{-0.5em}
\noindent{\bf Implementation details.}
We set $\lambda_c$, $\lambda_{gp}$, $\lambda_{AU}$, $\lambda_{g}$, $\lambda_{l}$, $\lambda_{adv}$ to be 1, 10, 250, 20, 20, 25, respectively. Our method is trained by an Adam optimizer with the learning rate 0.0001 and the exponential decay rates set to be $(\beta_1, \beta_2)=(0.5, 0.99)$. We evaluate the black-box attack performance of the models utilizing the \emph{attack success rate}(ASR) at FAR@0.01 and the confidence scores returned by commercial APIs.

\noindent{\bf Dataset.}
We train the model on two public datasets: 1) CelebA-HQ \cite{hq} is a high-quality face dataset, which contains 30,000 face images. 2) LFW \cite{lfw} is a challenging dataset that collects 13,233 images with complex environmental factors and is a common dataset for face recognition tasks. We remove the images whose AU confidence is below 95\% as extracted by Openface \cite{openface}, then randomly select 10\% of each dataset as the test set and the remaining images as the training set. Four pairs of images from CelebA \cite{celeba} with the same identity are used as attack targets for training and testing, respectively, since CelebA \cite{celeba} contains multiple images of one identity. Besides, the real state set in subsection \ref{ass} are obtained from the RaFD dataset \cite{rafd}.

\noindent{\bf Competitors.}
We compare our approach to the baselines PGD \cite{pgd}, MI-FGSM \cite{boosting}, SemanticAdv \cite{semanticadv}, TIP-IM \cite{zj21iccv} and AMT-GAN \cite{AMT}. Since our work belongs to the branch of GAN based unrestricted adversarial attack \cite{song18}, which is budget-free. Similar to \cite{song18}, we compare our method MAA/GMAA to both restricted and unrestricted adversarial methods w.r.t attack performance and visual naturalness. All restricted methods are setted to the size of perturbation $ \epsilon = 12 $. And all baselines are equipped with the transferability enhancement module for a fair comparison.

\noindent{\bf Target models.}
Following \cite{AMT}, we choose IR152 \cite{ir152}, IRSE50 \cite{irse50}, Facenet \cite{facenet} and Mobileface \cite{mb} as the attacked FR models, with three of them serving as white-box models during training and the remaining as the black-box model for testing.

\vspace{-0.5em}
\subsection{Comparison Study}
\vspace{-0.5em}
This subsection shows the comparison results of the MAA method and competitors in terms of attack performance and visual quality.

\noindent{\bf Comparison of attack performance on commercial API.} 
We evaluate the performance of each method against the commercial APIs Face++\footnote{\url{https://www.faceplusplus.com/}} and Tencent\footnote{\url{https://cloud.tencent.com/document/product/867/44987}}. Fig. \ref{api} exhibits the average confidence score of Face++ and Tencent between the adversarial example and the test image of the target identity. Our method MAA achieves the highest score, outperforming all competitors on both datasets, as shown in Fig. \ref{api}. Furthermore, in Fig. \ref{apiexp}, we display the confidence scores for five typical expressions (happy, angry, sad, surprised, and expressionless) to demonstrate that different AU vectors have little effect on attack performance and are more influenced by the dataset.

\noindent{\bf Comparison of black-box attack success rate.} 
Since our model can give us a semantic continuous adversarial manifold and we want to make sure the comparison is fair, we randomly sample the AU vector to get test adversarial examples for calculating the black-box ASR. Tab. \ref{tsingle} shows the black-box ASR of each method under four FR models and two datasets. Obviously, our method has good performance for black-box attacks, i.e., it has strong transferability.

\noindent{\bf Comparison of visual quality.}
We choose TIP-IM and AMT-GAN, two recent approaches with high black-box ASR, as benchmarks for our assessment of visual quality. Fig. \ref{vis} shows the adversarial examples of each method, and the target image is shown in Fig. \ref{structure} highlighted by a green square. In particular, to demonstrate that our method can generate semantically continuous adversarial examples, Fig. \ref{vis} displays the adversarial examples generated by MAA that continuously transform on four expressions (expressionless, disgusted, happy and surprised in succession). Note that although only 20 adversarial examples are presented in Fig. \ref{vis}, our method can generate an infinite number of adversarial examples by continuously interpolating between AU vectors since MAA establishes a correspondence with the AU vector space. Moreover, our method has a natural visual quality and is gender-insensitive.


\vspace{-0.5em}
\subsection{Attack state set}\label{ass}
\vspace{-0.5em}
\noindent{\bf Attack real state set.} 
To avoid serendipity, two FR models, three different test targets, and three adversarial attack methods were employed to assess the effectiveness of attacking state set on enhancing the adversarial examples' generalizability. Particularly, one of the FR models is Facenet with high accuracy, and the other is Mobileface with a lightweight network. Targets *, 1, 2 and 3 are shown in Fig. \ref{intro3}. The state set consists of several common expression states (angry, contemptuous, disgusted, fearful, happy, sad and surprised), and each correlates to an image. By comparing the results in Tab. \ref{rss}, we can summarize that the model trained to attack target * generalizes poorly to test targets 1, 2, 3, whereas adversarial examples generated on the state set generalize better to the test target, even though the test target is not used to train the model.

We try to add more images in the state set to further improve the performance. Tab. \ref{rss2} shows the ASR results that the target state set contains each expression state corresponding to 1/ 2/ 3 image(s), respectively. We demonstrate that the generalizability of the adversarial examples can be effectively strengthened as long as the state set contains a single image of each state, and additional images have minimal impact on the results.

\noindent{\bf Attack generated state set.} 
For face image datasets that do not contain rich states, we generate the target state set by the pre-trained generator $G_{exp}$ with AU vectors of common expressions. Tab. \ref{gss} shows the comparison results of training on a single target and a generated state set, demonstrating that training the model on the target state set enhances the generalization ability of adversarial examples.
\vspace{-0.2em}
\section{Future Work}
\vspace{-0.2em}
 Considering the ubiquitous applicability of expressions in the facial adversarial attacks, the expression state space was chosen to implement GMAA in this paper. We employ the FACS as the prior domain knowledge to implement GMAA, while the paradigm GMAA can be broadly generalized by integrating other domain information, such as illumination, posture, etc. By selecting different state spaces, our work can be generalized to other adversarial attacks with more general image categories.
\vspace{-0.2em}
\section{Conclusion}
\vspace{-0.2em}
In this paper, we provide a novel paradigm GMAA that broadens both target domain and adversarial domain to enhance the performance of adversarial attack. For the target domain, GMAA optimizes generalization to the target identity by attacking the state set instead of a single image. Additionally, GMAA leverages the domain knowledge to expand adversarial domain from discrete points to semantic continuous manifold. Numerous comparative experiments have verified that GMAA has a better attack performance and a more natural visual quality than other competitors. Moreover, the generalized attack module can be extended to a wide of applications.

\vspace{-0.2em}
\section{Acknowledgement}
\vspace{-0.2em}
This work was supported and partially funded by the National Natural Science Foundation of China (Grant No. 62106116). This work was also supported in part by ZJNSFC under Grant LQ23F010008. We also would like to thank Han Fang of NUS for the valuable discussion.

\clearpage

{\small
\bibliographystyle{ieee_fullname}
\bibliography{egbib}
}

\onecolumn

\vspace{-0.2em}
\section{Ablation studies}
\begin{figure}[h]
\centering
\includegraphics[width=1\textwidth]{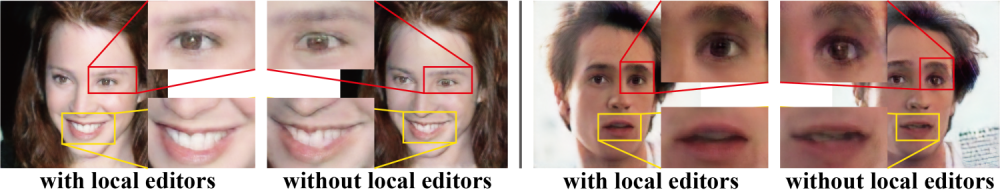}
\caption{Ablation studies of three local editors.}
\label{abl}
\end{figure}

\vspace{-0.2em}
We verify the validity of the three local editors and $D_{AU}$. 
Fig. \ref{abl} demonstrates the poor visual quality of the adversarial examples without three local editors. Tab. \ref{T1} illustrates that the $D_{AU}$ and local editors help to improve the expression accuracy of the generated adversarial examples.

\begin{table}[htbp]
\centering
\caption{MSE between the AU vector of the generated adversarial examples (detected by OpenFace) and the given AU vector.}
\setlength{\tabcolsep}{13mm}{
\begin{tabular}{cccc}
\toprule
& Without $D_{AU}$& Without local editors& Original\\
\midrule
MSE&0.5549&0.6283&\textbf{0.3582}\\
\bottomrule
\label{T1}
\end{tabular}}
\end{table}

\newpage
\vspace{-0.2em}
\section{More visualization results}
\vspace{-0.2em}

Due to the limited length of the main text, we have attached more visualization results here, which show that our model MAA outperforms the comparative models AMT-GAN and TIP-IM.

\begin{figure}[htbp]
  \centering
  \begin{subfigure}{1\textwidth}
    \includegraphics[width=1\linewidth]{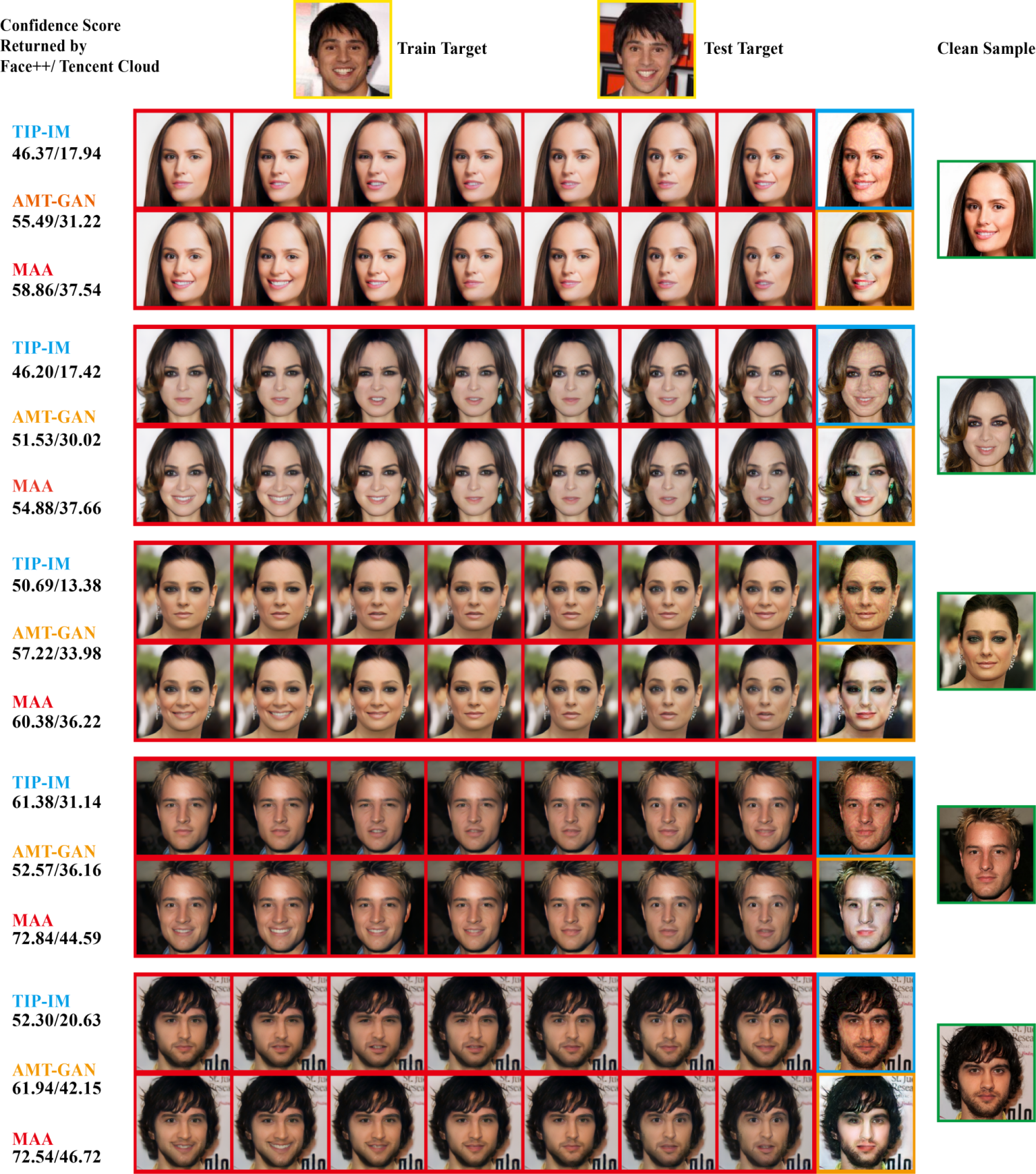}
  \end{subfigure}
  \end{figure}
  \addtocounter{figure}{-1}
  \begin{figure}
  \begin{subfigure}{1\textwidth}
    \includegraphics[width=1\linewidth]{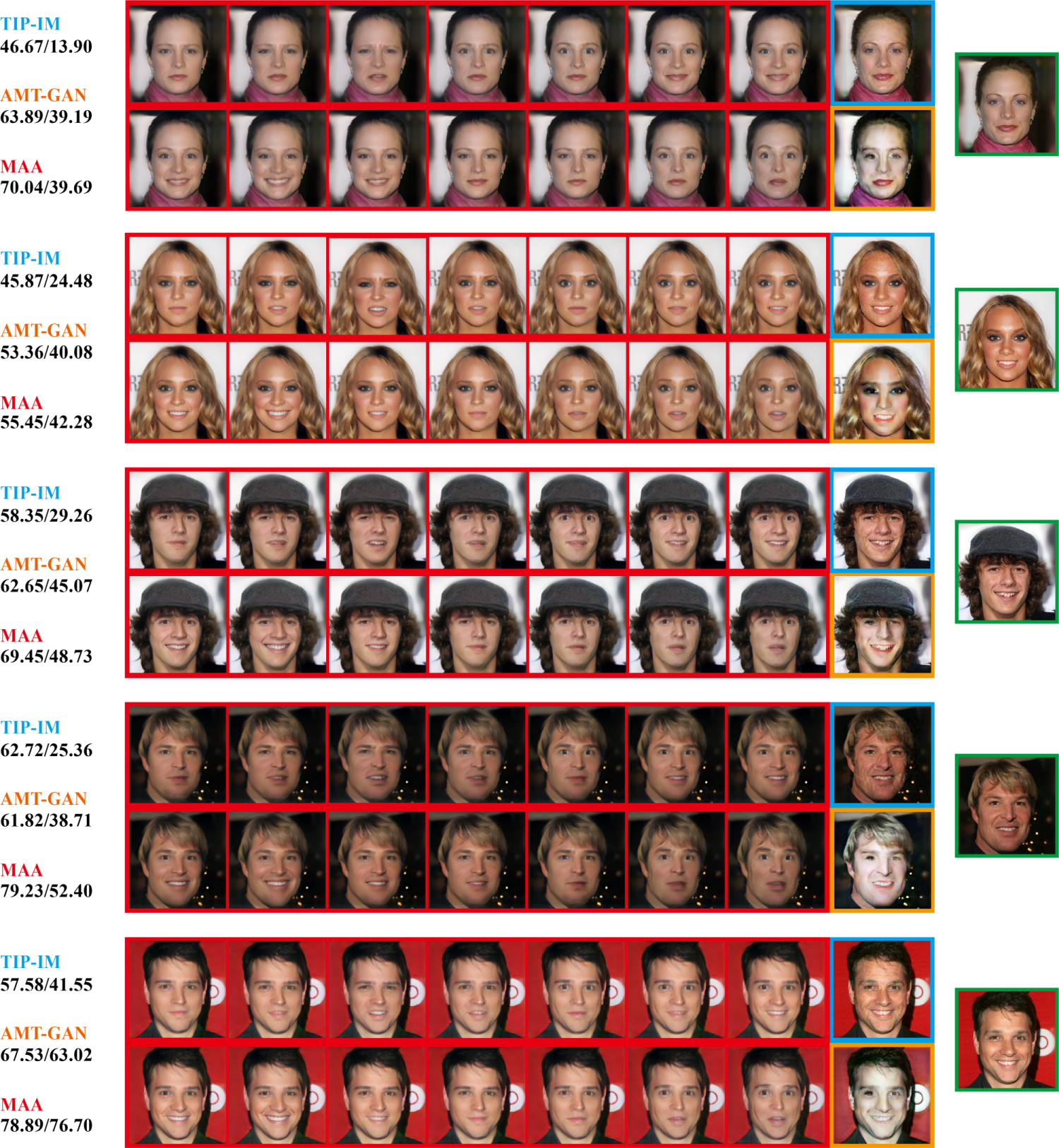}
  \end{subfigure}
  \caption{Visualization results of TIP-IM (highlighted by blue boxes), AMT-GAN (highlighted by orange boxes) and MAA (highlighted by red boxes). All the methods are trained to attack train target and the results returned by Face++/Tencent are the confidence score between the test target and adversarial examples. The results of MAA are the average scores of adversarial examples based on the same clean sample.}
  \label{vis-s}
\end{figure}

\end{document}